%% file: acl_latex_arr_jan26.tex
\pdfoutput=1

\documentclass[11pt]{article}
\usepackage[table,dvipsnames]{xcolor}
\usepackage[preprint]{acl}

\usepackage{times}
\usepackage{latexsym}

\usepackage[T1]{fontenc}

\usepackage[utf8]{inputenc}

\usepackage{microtype}

\usepackage{inconsolata}

\usepackage{graphicx}

\input{math_commands.tex}

\usepackage{hyperref}
\usepackage{url}

\usepackage{graphicx}
\usepackage{caption}
\usepackage{multirow}
\usepackage{color}
\usepackage{tabularx}
\usepackage{booktabs}
\usepackage{subcaption}
\usepackage{enumitem}
\usepackage{listings}
\usepackage{wrapfig}
\usepackage{makecell}

\title{ToolRM: Outcome Reward Models for \\ Tool-Calling Large Language Models}

\author{
    Mayank Agarwal,
    Ibrahim Abdelaziz,
    Kinjal Basu, \\
    {\bf 
    Merve Unuvar,
    Luis A. Lastras,
    Yara Rizk,
    Pavan Kapanipathi
    } \\
    IBM Research, USA \\
  \normalsize{\{mayank.agarwal, ibrahim.abdelaziz1, kinjal.basu\}@ibm.com}
}

\newcommand{\fcrewardbench}{FC-RewardBench}
\newcommand{\modelName}{ToolRM}%
\newcommand{\bon}{Best-of-$n$}

\begin{document}

\maketitle
\begin{abstract}
As large language models (LLMs) increasingly interact with external tools, reward modeling for tool use has emerged as a critical yet underexplored area of research. Existing reward models, trained primarily on natural language outputs, struggle to evaluate tool-based reasoning and execution. To quantify this gap, we introduce {\fcrewardbench}, the first benchmark to systematically evaluate reward models in tool-calling scenarios. Our analysis shows that current reward models frequently miss key signals of effective tool use, highlighting the need for domain-specific modeling. We address this by proposing a training framework for outcome reward models using data synthesized from permissively licensed, open-weight LLMs. 
We introduce {\modelName} -- a suite of reward models for tool-use ranging from 1.7B to 14B parameters.
Across diverse settings, these models consistently outperform general-purpose baselines. Notably, they achieve up to a 25\% improvement with {\bon} sampling, while also improving robustness to input noise, enabling effective data filtering, and supporting RL-training of policy models.
\end{abstract}

\section{Introduction}
\input{sections_acl/introduction}

\section{Related Work}

\input{sections_acl/related_work}

\section{Methodology}
\input{sections_acl/orm}

\section{Experimental Setup}
\label{sec:experimental-setup}
\input{sections_acl/experimental_setup}

\section{Results}
\input{sections_acl/results}

\section{Conclusion}
\input{sections_acl/conclusion}

\bibliography{iclr2025_conference}

\newpage
\appendix

\section{Appendix}
\label{sec:appendix}
\input{sections_acl/appendix}

\end{document}

%% file: math_commands.tex
\usepackage{amsmath,amsfonts,bm}

\def\eqref#1{equation~\ref{#1}}

\def\1{\bm{1}}

\DeclareMathAlphabet{\mathsfit}{\encodingdefault}{\sfdefault}{m}{sl}
\SetMathAlphabet{\mathsfit}{bold}{\encodingdefault}{\sfdefault}{bx}{n}

%% file: sections_acl/introduction.tex
Large language models (LLMs) 
have rapidly advanced the field of artificial intelligence (AI), achieving strong performance across a wide range of tasks, including complex question answering, code generation, and multi-step reasoning \citep{li2025system}. As these models are increasingly deployed in real-world systems, the need for them to interact with external tools has become critical.
Tool calling enables LLMs to invoke external functions such as APIs, databases, calculators, and search engines \citep{prabhakar2025apigen, zhang2024xlam, abdelaziz2024granite, liu2024toolace, lin2024hammer}, shifting their role from standalone text generators to orchestrators of complex workflows. This capability underpins their application in autonomous agents, virtual assistants, and multimodal systems.

Training these LLMs effectively requires reward models, which are integrated into the learning pipeline through reinforcement learning (RL), preference optimization \citep{wang2023aligning}, and rejection sampling fine-tuning \citep{touvron2023llama, qwen2.5}.
Reward models provide learned signals that estimate output quality, enabling scalable evaluation without requiring human judgment on every example. Broadly, they fall into two categories: process reward models (PRMs) \citep{lightman2023let}, which score intermediate reasoning steps, and outcome reward models (ORMs) \citep{cobbe2021training}, which evaluate only the final answer. 

\begin{figure*}
    \centering
    \includegraphics[width=\linewidth]{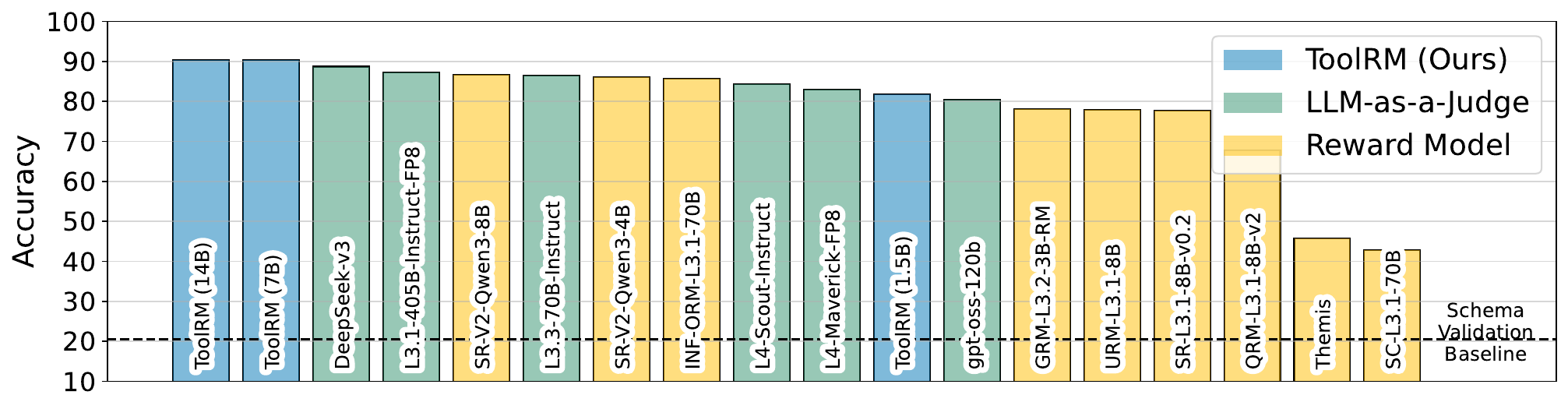}
    \caption{Performance of {\modelName}, top reward models from RewardBench, Tool-augmented RM (Themis), and leading LLMs-as-judges on {\fcrewardbench}. \textit{Note:} Model names are abbreviated for conciseness (e.g., L3.1-xx, SR-xx, and SC-xx correspond to Llama-3.1-xx, SkyWorks-Reward-xx, and SkyWorks-Critics-xx, respectively). Full model names are provided in Appendix \ref{apdx:fcrb-experiment-details}.}
    \label{fc_reward_bench_eval}
\end{figure*}

Despite their successes, current reward models are designed primarily for natural language outputs \citep{zhong2025comprehensive}. Reward modeling for tool calling remains an underexplored area, with two notable gaps: (a) no dedicated benchmark exists for evaluating reward models in the function-calling domain\footnote{Tool-use, tool-calling, and function-calling are used interchangeably throughout the paper}, and (b) existing reward models fail to capture the nuances of tool-based reasoning and execution. In order to address these gaps, we first introduce {\fcrewardbench} -- a comprehensive benchmark specifically designed to evaluate reward models on tool-calling tasks.
Derived from the Berkeley Function Calling Leaderboard (BFCL) Version 3 \citep{patilberkeley}, the dataset contains 1500 user inputs paired with correct and incorrect function calls.
We benchmark several state-of-the-art general-purpose reward models on {\fcrewardbench}, and our analysis (Figure \ref{fc_reward_bench_eval}) shows that these models often fail to capture key aspects of successful tool use, hence failing to capture the nuances of tool-based reasoning and execution. 
To this end, next, we introduce {\modelName}, a collection of specialized ORMs for tool calling. Trained on preference data synthesized from a diverse set of open-source function-calling models, {\modelName} outperforms much larger reward models and LLMs-as-Judges on {\fcrewardbench}. In downstream applications, {\modelName} achieves up to 25\% average improvement across multiple benchmarks in a {\bon} setting, enables effective data filtering that produces stronger fine-tuned models with significantly less training data, and enable RL-training of policy models without requiring ground-truth labels for reward computation.

In summary, our contributions are:

\begin{itemize}
    \item We introduce {\fcrewardbench} ($\S$\ref{sec:fc-reward-bench-eval-dataset}), the first benchmark for evaluating reward models for tool-calling task, and demonstrate that existing RMs struggle in this domain ($\S$\ref{sec:fc-reward-bench-results}).
    
    \item We propose a framework for training outcome reward models for tool-calling tasks using data generated from permissively licensed, open-weight LLMs, and train a suite of reward models ({\modelName}-1.7B, 7B, 14B) ($\S$\ref{sec:experimental-setup}).
    
    \item We show that {\modelName} deliver substantial practical gains:  
    \quad (1) up to 25\% improvement when used for {\bon} sampling ($\S$\ref{sec:bon-sampling-results});  
    \quad (2) superior fine-tuned models using only 50\% of the data after {\modelName}-based filtering ($\S$\ref{sec:data-filtering}); and  
    \quad (3) RL-trained policies using {\modelName} rewards match the performance of models trained with ground-truth–dependent rewards ($\S$\ref{sec:policy-optimization}).
\end{itemize}

%% file: sections_acl/related_work.tex
\subsection{Tool Calling}
Tool calling has extended LLMs beyond static knowledge to tasks requiring external retrieval \citep{schick2023toolformer}, reasoning \citep{he2023solving}, orchestration \citep{jain2024smartflow}, and code execution \citep{gao2023pal}. Early prompting-based approaches such as ReAct \citep{yao2023react} inspired refinements for efficiency \citep{xu2023rewoo}, performance \citep{shinn2023reflexion, yang2023mm}, or balanced trade-offs \citep{crouse2023formally}. Recent models now provide built-in tool use \citep{reid2024gemini, codegemma_2024, cohereforai-c4ai-command-r, llama-3, jiang2023mistral} or are fine-tuned for this capability \citep{qin2023toolllm, tang2023toolalpaca, patil2023gorilla, abdelaziz2024granite}. To assess and enhance these capabilities, benchmarks \citep{guo2024stabletoolbench, patil2023gorilla}, curated datasets \citep{liu2024toolace, qian2025smart}, and autonomous tool construction methods \citep{qian2023toolink, qian2023creator} have been proposed. 

\subsection{RL for Tool-Use Alignment}

Reinforcement Learning has become a powerful approach for aligning LLMs with effective tool use. Search-R1 \citep{jin2025search} trains LLMs to iteratively refine search queries, showing RL feedback balances exploration and retrieval precision. ToRL \citep{li2025torl} enables autonomous discovery of tool-use strategies, with rewards driving emergent behaviors like strategic invocation and adaptive reasoning. ReTool \citep{feng2025retool} interleaves code execution with natural language reasoning, using outcome feedback to guide tool invocation, improving mathematical problem solving. Several works focus on reward design: ToolRL \citep{qian2025toolrlrewardtoollearning} studies how reward type, granularity, and temporal dynamics affect alignment; StepTool \citep{yu2024steptool} uses step-level reward shaping and policy-gradient optimization for multi-step tasks; CodeTool \citep{lu2025codetool} combines RL with step-level supervision to encourage intermediate reasoning; SWE-RL \citep{wei2025swe} leverages software evolution data to optimize reasoning over action sequences, capturing temporal dependencies; and iTool \citep{zeng2025itoolreinforcedfinetuningdynamic} mitigates performance decay from synthetic data via iterative reinforced fine-tuning with Monte Carlo Tree Search. 
Together, these works show RL’s effectiveness in aligning LLMs for general-purpose tool use, though none explicitly employ an ORM that evaluates or optimizes entire sequences of tool interactions.

\subsection{Reward Modeling}

Reward models (RMs) provide scalar preference signals that guide LLMs during preference optimization or RL \citep{wang2024secrets}. RMs can be broadly classified as Outcome Reward Models (ORMs), which score only the final output, and Process Reward Models (PRMs), which evaluate intermediate reasoning steps \citep{zhong2025comprehensive}. Early verifier-based methods in math \citep{cobbe2021training} established ORMs, while later work contrasted outcome- vs. process-based supervision \citep{uesato2022solvingmathwordproblems} and developed PRMs that reward coherent stepwise reasoning \citep{lightman2023let}. However, PRMs face robustness and supervision challenges \citep{zhang2025lessons}, as evidenced by failed attempts reported by \citet{guo2025deepseek}. ORMs, in contrast, have scaled more reliably \citep{lin2025exploringlimitoutcome}, with advances like Skywork-Reward \citep{liu2024skywork} achieving state-of-the-art results on RewardBench \citep{lambert2024rewardbench}. More recently, tool-augmented reward models \citep{li2024toolaugmented} allow RMs to use external tools for more accurate preference scoring. 
While prior work has focused on free-text, math, and code domains, to our knowledge this is the first to introduce ORMs for tool calling, where outcomes are defined by sequences of tool calls.

%% file: sections_acl/orm.tex
\subsection{\fcrewardbench\ Evaluation Dataset}
\label{sec:fc-reward-bench-eval-dataset}

While several benchmarks evaluate RMs on tasks involving chat, reasoning, safety \citep{lambert2024rewardbench}; factuality, instruction following, and math \citep{malik2025rewardbench}, there remains a notable gap in the evaluation of RMs for function-calling tasks. 
To bridge this gap, we propose \fcrewardbench, a benchmark specifically designed to evaluate RMs on function-calling tasks. This dataset comprises 1500 unique data points, each containing a user query, a tool catalog (tools available to the model to answer the user query), and the associated correct and incorrect tool calls for a given user query.

To construct \fcrewardbench, we utilize the single-turn splits of the BFCL-v3 dataset \citep{patilberkeley}. The tool catalog, user query, and the correct tool calls in the dataset are directly sourced from BFCL-v3. Incorrect tool calls are generated using a pool of 25 language models, spanning sizes from 0.5B to 685B parameters. Each model is prompted to generate a tool call in response to the user query. The outputs are compared against the ground-truth, and only the incorrect generations are retained. From this pool, we randomly sample one incorrect call per instance to prevent over-representation from any single user query. Finally, 1,500 such examples are randomly selected to form the final dataset.

\begin{table}
    \centering
    \resizebox{0.75\linewidth}{!}{%
    \begin{tabular}{c c}
        \toprule
       \textbf{Error Type}  & \textbf{Count} \\
       \midrule
       
       Incorrect Parameter Value  & 650 \\
       Incorrect Function Name  &    403 \\
       Incorrect number of functions    &   245 \\
       Missing Optional Parameter   &   78  \\
       Missing Required Parameter   &   45  \\
       Incorrect Parameter Type  &   43 \\
       Unexpected Parameter  &  21  \\
       Incorrect output format  &   15 \\
       \bottomrule
    \end{tabular}
    }
    \caption{Breakdown of errors in the \fcrewardbench\ dataset. The majority of errors in the dataset are subtle and hard to identify.}
    \label{tab:fc-reward-bench-errortypes}
\end{table}

Table~\ref{tab:fc-reward-bench-errortypes} presents a breakdown of error types observed in the dataset. Notably, a majority of the incorrect calls involve subtle errors such as incorrect parameter values, missing optional parameters, or an incorrect number of functions, which are non-trivial to detect. These characteristics require the RM to demonstrate a deeper understanding of the function-calling task, making \fcrewardbench\ a challenging and discriminative benchmark. 
Figure~\ref{fig:fc-reward-bench-example} shows a representative example from the dataset, and additional details about the benchmark are provided in Appendix \ref{apdx:fcrb-benchmark-details}.

\subsection{Reward Modeling}

For pairwise preference modeling, RMs are commonly formulated using the Bradley–Terry model \citep{bradley1952rank}, which defines the probability that output $y_{+}$ is preferred over $y_{-}$ given an input $x$ as:

\begingroup
\setlength{\abovedisplayskip}{0pt}
\setlength{\abovedisplayshortskip}{0pt}
\setlength{\belowdisplayskip}{5pt}
\setlength{\belowdisplayshortskip}{0pt}

\begin{multline}
p(y_{+} \succ y_{-} \mid x) =
\frac{e^{(r(x, y_{+}))}}{e^{(r(x, y_{+}))} + e^{(r(x, y_{-}))}} 
\\ =  \sigma( r(x, y_{+}) - r(x, y_{-}))
\end{multline}
\endgroup

where $r(x,y)$ is a scalar reward function, and $\sigma$ is the sigmoid function.

Training requires curating a dataset of pairwise preferences $D = \{(x, y_{+}, y_{-}) : y_+ \succ y_{-}\}$, with preferences obtained through either human annotations \citep{stiennon2020learning, ouyang2022training} or synthetic generation methods \citep{pace2024west, Hosseini2024VSTaRTV}. The reward function $r$ is parameterized by a neural network $r_\theta$, typically initialized from a supervised fine-tuned model with the final layer replaced by a linear head.

The parameters of $r_\theta$ are estimated from the dataset $D$ using maximum likelihood estimation of the following objective:

\begingroup
\setlength{\abovedisplayskip}{0pt}
\setlength{\abovedisplayshortskip}{0pt}
\setlength{\belowdisplayskip}{0pt}
\setlength{\belowdisplayshortskip}{0pt}

\begin{multline}
    J(r) = \max_{r_\theta} \mathbb{E}_{(x, y_+, y_-) \sim D} [ \\
    log(\sigma(r_\theta(x, y_+) - r_\theta(x, y_{-}))]
\end{multline}
\endgroup

In this work, we use reward centering \citep{eisenstein2023helping} to ensure that rewards are zero-centered. This is achieved by adding the following regularization term to the optimization objective:

\begingroup
\setlength{\abovedisplayskip}{0pt}
\setlength{\abovedisplayshortskip}{0pt}
\setlength{\belowdisplayskip}{0pt}
\setlength{\belowdisplayshortskip}{0pt}

\begin{multline}
    J_{reg}(r) = J(r) + \eta \mathbb{E}_{(x, y_+, y_{-}) \sim D} [ \\
    (r_{\theta}(x, y_+) + r_{\theta}(x, y_{-}))^2]
\end{multline}
\endgroup

where $\eta$ is a small positive value hyperparameter. 

\subsection{{\modelName} Training Data Generation}

To train ORMs for function-calling tasks, we require data consisting of user queries, tool catalogs, and the corresponding correct and incorrect tool calls.
We construct this data by leveraging a diverse set of open-source, permissively licensed language models with function-calling capabilities. Specifically, we use publicly available function-calling datasets, which provide user queries, tool catalogs, and ground-truth tool call sequences. For each query, we prompt the models to generate tool calls using the tools specified in the dataset.

The generated tool calls are then compared against the ground-truth sequences. Outputs that deviate from the ground truth are retained as incorrect examples, while matching outputs are discarded. %
This procedure enables the collection of data that reflects the natural variability and error patterns of real-world models. It captures not only common mistakes but also subtle and complex failure modes that are difficult to anticipate or enumerate manually.

%% file: sections_acl/experimental_setup.tex
\paragraph{Training Data:} To create training data for the RM, we select open-source datasets that cover various aspects of function-calling, such as the API-Gen dataset \citep{liu2024apigen} for single-turn interactions, the Schema-Guided Dialogue (SGD) dataset \citep{rastogi2020towards} for multi-turn interactions with tool invocations and responses, and the xlam-irrelevance\footnote{\url{https://huggingface.co/datasets/MadeAgents/xlam-irrelevance-7.5k}} dataset for cases where the system lacks sufficient information to respond to a user query.

Since these datasets are common training datasets and our primary focus is to elicit representative incorrect behavior from the model, we follow \citet{lin2024hammer} and obfuscate the data samples to avoid the model regurgitating its training data. We obfuscate the samples by replacing function and parameter names with randomly generated strings and reordering the keys in the function schema. 

We then use a collection of 11 permissively-licensed, open-weight models to generate the training data. The pool includes both general-purpose instruction-tuned models with function-calling capabilities and function-calling specific models, with parameter counts ranging from 0.5B to 32B. Specifically, we use the Qwen2.5-Instruct \citep{qwen2.5} and Granite 3.3-Instruct \citep{granite2024granite} model series, along with Granite-20b-function-calling \citep{abdelaziz2024granite}, SmolLM2 \citep{allal2025smollm2}, Mistral-7b-Instruct-v0.3
and Mistral-Nemo-Instruct-2407.

After generating outputs from the model pool and keeping only the incorrect ones, we subsample one incorrect output per input user query to prevent over-representation from a user query in the training data. Overall, this results in 180K training data samples divided into 85K single and multi-turn data each, and 10K irrelevance data. 
The full list of models used to generate the training data, along with a few training data samples, is provided in Appendix \ref{apdx:training-datagen-details}.

\vspace{-5pt}
\paragraph{Model architecture:} We use the Qwen-2.5-Instruct (1.5B, 7B, and 14B parameter variants) models \citep{qwen2.5, qwen2} as the base architecture for our RMs. 
We initialize the RMs with the instruction-tuned model weights and replace the final language modeling head with a linear layer that maps the hidden representation to a scalar reward value.

The RMs accept the specifications of available functions, conversation history, and the generated tool call as input and produce a scalar reward as output (refer to Appendix \ref{apdx:toolrm-prompt} for prompt template). We train all RMs for 1 epoch with a learning rate set to 1e-6, a cosine learning rate schedule with warmup set to 3\% of total steps, and the reward centering coefficient set to 0.01.

\vspace{-5pt}
\paragraph{Benchmarks:} In addition to {\fcrewardbench}, we evaluate models on the following commonly used function-calling benchmarks: Berkeley Function Calling Leaderboard (BFCL) v3 \citep{patilberkeley}, API-Bank \citep{li2023apibank}, ToolAlpaca \citep{tang2023toolalpaca}, NexusRaven API Evaluation \footnote{\url{https://huggingface.co/datasets/Nexusflow/NexusRaven_API_evaluation}}, and SealTools \citep{wu2024sealtoolsselfinstructtoollearning}. For API-Bank, we evaluate on the Call (API-Bank-1) and Retrieval+Call (API-Bank-2) splits. Table \ref{tab:eval-benchmarks} summarizes their key statistics and characteristics. We highlight that these benchmarks vary in difficulty, encompassing single and multi-turn queries, nested tool calls, and evaluation sets collected from both real users and synthetically generated.

\vspace{-5pt}
\paragraph{Baselines:} 
To evaluate performance on {\fcrewardbench}, we select eight RMs from \textit{RewardBench}, spanning sizes from 3B to 70B parameters. We chose models that achieved high scores on RewardBench and support tool use in their chat template, which helps mitigate performance degradation due to prompt variability.
In addition to these specialized RMs, we include six LLMs as judges, ranging from 70B to 685B parameters. See Appendix \ref{apdx:fcrb-experiment-details} for the complete list of models. 

For downstream task evaluations, we select the strongest function-calling models -- the xLAM-2 series \citep{prabhakar2025apigenmtagenticpipelinemultiturn} -- and the strongest generic instruction-tuned models -- the Qwen3 series \citep{yang2025qwen3} -- from the BFCL-v3 leaderboard. Both of these model series cover a wide range of sizes (0.6B to 70B), enabling a comprehensive assessment of {\modelName} across model scales.

%% file: sections_acl/results.tex
We evaluate our proposed RM to answer the following three research questions (RQ): \\
\textbf{RQ1:} How does {\modelName} compare to existing RMs on {\fcrewardbench}? \\
\textbf{RQ2:} Can {\modelName} improve the performance during inference through {\bon} sampling? And, \\
\textbf{RQ3:} Can {\modelName} lead to better models through reward-guided data filtering or policy optimization?

\subsection{RQ1: {\fcrewardbench} evaluation}
\label{sec:fc-reward-bench-results}

We evaluate {\modelName} against state-of-the-art RMs from RewardBench \citep{lambert2024rewardbench}, Tool-Augmented RM (Themis) \citep{li2024toolaugmented}, as well as leading LLMs used in an LLM-as-a-Judge setting, on the {\fcrewardbench} dataset.

RMs are evaluated by comparing scores assigned to the correct tool call outputs and incorrect tool call outputs for the same input.
A prediction is counted as correct when the score for the correct tool call exceeds that of the incorrect one. LLMs-as-Judges are evaluated with a pairwise comparison prompt, where both candidate tool calls are presented and the model is instructed to select the correct one. To avoid position bias, the order of candidates is randomized. Experimental details, including the full prompt template, are provided in Appendix \ref{apdx:fcrb-experiment-details}.
We show the results in Figure \ref{fc_reward_bench_eval} and observe the following:

\begin{itemize}[topsep=0pt, parsep=0pt, leftmargin=*]
    \item \textbf{Existing reward models struggle with tool-calling tasks.} 
    Specialized RMs underperform and often fail to generalize to tool-calling behaviors -- e.g., the Tool-Augmented RM (Themis) achieves only 45\% accuracy on \fcrewardbench, and rule-based method fares even worse. 
    While LLMs-as-Judges achieve strong accuracy (exceeding 80\%), their large parameter counts make them computationally expensive.

    \item \textbf{{\modelName} achieves state-of-the-art accuracy with high efficiency.} 
    {\modelName}-7B and {\modelName}-14B outperform all other generative and sequential classifier models, and the {\modelName}-1.5B variant even surpasses the gpt-oss-120B model, approaching the performance of much larger Llama-4 models.
\end{itemize}

The primary purpose of \fcrewardbench\ is to enable quick evaluation of RMs without requiring costly downstream experiments. Therefore, performance on \fcrewardbench\ should correlate strongly with downstream results. In Appendix \ref{apdx:fcrb-correlation}, we report correlations between 11 RMs on \fcrewardbench\ and their \bon scores on five downstream benchmarks. With an average correlation of 0.84, \fcrewardbench\ provides a reliable and efficient proxy for downstream RM evaluation.

\subsection{RQ2: {\bon} sampling with {\modelName}}
\label{sec:bon-sampling-results}

In this section, we evaluate {\modelName} in a {\bon} setting across multiple generator models. For each input, we sample $n=32$ independent generations using temperature $T=0.6$ from the generator model and use {\modelName} to score and select the highest-ranked generation as the final output. Intuitively, a stronger RM should more reliably identify the correct tool call, thereby improving task performance. We compare against three baselines: 1) Greedy Decoding, 2) Majority Voting -- where the most frequently occurring final answer is selected as the output, and 3) Schema Validation -- where we compare the output against the input tool schema and return the generation with the highest likelihood that validates the schema.
For non-BFCL benchmarks, we report the Full Sequence Matching metric \citep{nestful}, which checks whether the predicted tool sequence -- including tool names and argument-value pairs -- exactly matches the gold sequence. For BFCL, we use its native evaluation metrics: AST-based scores for single-turn tasks and state-based/response-based metrics for multi-turn cases.

\begin{figure*}[h]
    \centering
    \includegraphics[width=\textwidth]{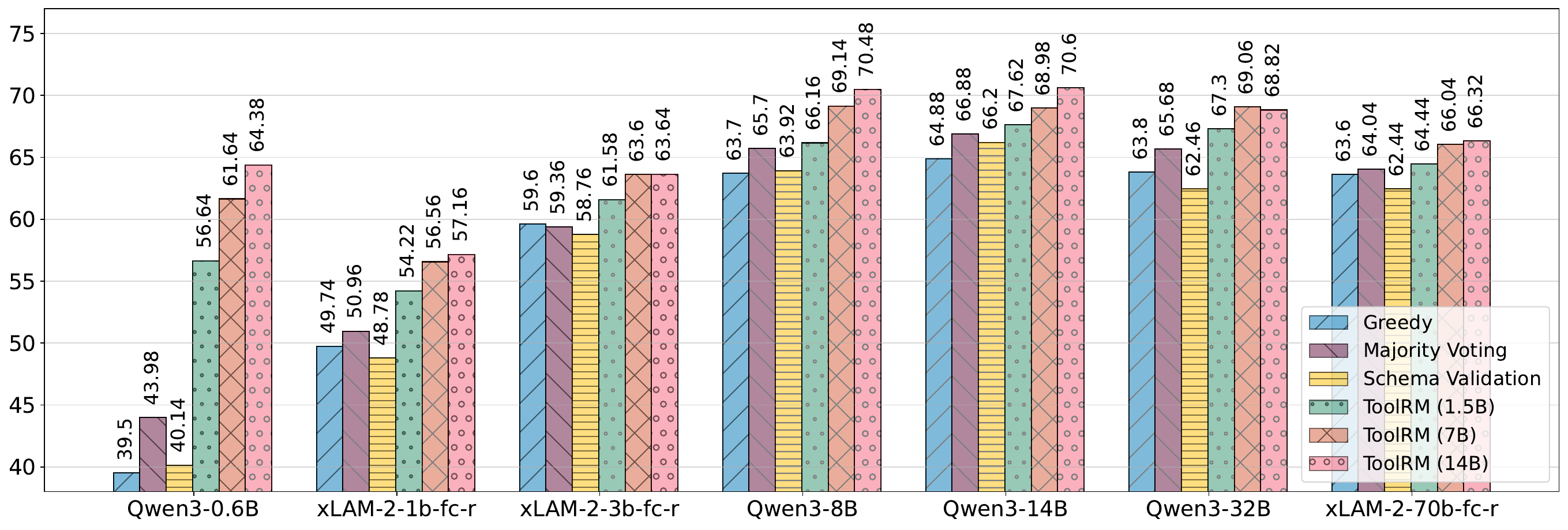}
    \caption{Performance of the Qwen3 series and xLAM-2 series in the {\bon} $(n=32)$ setting across five benchmarks: API-Bank-1, API-Bank-2, NexusRaven, ToolAlpaca, and SealTools.}
    \label{ood_fig}
\end{figure*}

Figure \ref{ood_fig} reports average performance across five benchmarks (API-Bank-1, API-Bank-2, ToolAlpaca, NexusRaven, and SealTools), while Table \ref{bfcl_v3_res} presents results on the BFCL-v3 dataset. We summarize the key insights below.

\begin{table*}[h]
\centering
\resizebox{0.8\textwidth}{!}{%
\begin{tabular}{cccccc}
\toprule
Model                                 & RM                                & Overall Acc    & Non-Live AST   & Live AST       & Multi-Turn Acc \\ 
\midrule
\multirow{4}{*}{Qwen3-1.7b}           & Greedy                            & 55.74          & 80.23          & 71.35          & 10.25          \\
                                      & Majority Voting                   & 57.96          & 83.90          & 74.24          & 10.12          \\
                                      & Schema Validation                  & 56.34          & 83.48          & 72.46          & 8.25           \\
                                      & {\modelName}-14B & \textbf{61.05} & \textbf{89.79} & \textbf{80.01} & \textbf{14.12} \\ \midrule
\multirow{4}{*}{Qwen3-8b}             & Greedy                            & 64.65          & 88.90          & 80.09          & 26.38          \\
                                      & Majority Voting                   & \textbf{67.96} & 90.33          & 81.72          & \textbf{33.13} \\
                                      & Schema Validation                  & 67.21          & 90.58          & 81.05          & 32.50          \\
                                      & {\modelName}-14B & 67.14          & \textbf{92.19} & \textbf{82.98} & 31.50          \\ \midrule
\multirow{4}{*}{xLAM-2-1b-fc-r}       & Greedy                            & 54.09          & 68.98          & 54.77          & \textbf{35.12} \\
                                      & Majority Voting                   & 53.51          & 69.42          & 54.92          & 31.50          \\
                                      & Schema Validation                  & 54.21          & 70             & 55.51          & 33.88          \\
                                      & {\modelName}-14B & \textbf{57.28} & \textbf{75.50} & \textbf{60.92} & 34.25          \\ \midrule
\multirow{4}{*}{Llama-xLAM-2-8b-fc-r} & Greedy                            & 71.14          & 84.31          & 67.80          & 67             \\
                                      & Majority Voting                   & 72.39          & 84.90          & 67.75          & \textbf{67.75} \\
                                      & Schema Validation                  & 70.89          & 84.79          & 66.47          & 65.38          \\
                                      & {\modelName}-14B & \textbf{72.52} & \textbf{87.73} & \textbf{72.46} & 61.62          \\ \midrule
\end{tabular}%
}
\caption{Performance of the Qwen3 and xLAM-2 series of models in the {\bon} $(n=32)$ setting on BFCL-v3.}
\label{bfcl_v3_res}
\vspace*{-5pt}
\end{table*}

\begin{itemize}[topsep=0pt, parsep=0pt, leftmargin=*]
    \item 
    \textbf{Small Language Models (SLMs) benefit most and can match or surpass larger models:} {\bon} sampling with {\modelName}-14B yields the largest gains for small generators. Qwen3-0.6B improves from 39.5\% to 64.4\% -- a gain of 24.9 points that surpasses Qwen3-32B (63.8\%) and xLAM-2-70B (63.6\%) with greedy decoding. Similarly, Qwen3-8B reaches 70.5\%, exceeding all greedy baselines by 5.6 points. On BFCL-v3, Qwen3-1.7B achieves improvements of 5.3 points on overall accuracy and 9.6 and 8.7 points on Non-Live AST and Live AST metrics respectively (Table \ref{bfcl_v3_res}).
    
    \item 
    \textbf{Diminishing returns for very large models:} Improvements for 32B+ generators are modest -- Llama-xLAM-2-32B-fc-r gains 2.1 points on non-BFCL benchmarks and 2.5 points on BFCL Live AST -- suggesting limited additional utility of {\bon} sampling with already-strong base models.
\end{itemize}

We also look at the breakdown of errors with greedy decoding and {\bon} sampling with {\modelName}-14B, and present the results in Appendix \ref{apdx:bon-analysis}.

\vspace{-5pt}
\paragraph{
{\bon} sampling improves model robustness:}
We examine the impact of {\bon} sampling on model robustness to noise in the input. We utilize RoTBench \citep{ye2024rotbench}, which comprises of 568 tool specifications and 105 user queries paired with tools with varying levels of noise. The \textit{Clean} split contains tool and parameter names that clearly reflect their usage, while the \textit{Slight}, \textit{Medium}, and \textit{Heavy} splits introduce increasing noise through operations such as character insertion and deletion, name reversal, and name swapping. The \textit{Union} split combines all noisy variants and represents the most challenging setting. Model performance is evaluated across three tasks: Tool Selection, Parameter Identification, and Content Filling.

\begin{table*}
\vspace*{-5pt}
\resizebox{\textwidth}{!}{%
\centering
\scriptsize
\setlength{\tabcolsep}{4pt} %
\renewcommand{\arraystretch}{1.2}
\begin{tabular}{c|cccc|cccc|cccc}
\toprule
\multirow{2}{*}{\textbf{\makecell{Generator \\ Model}}} 
& \multicolumn{4}{c|}{\textbf{Tool Selection}} 
& \multicolumn{4}{c|}{\textbf{Parameter Identification}} 
& \multicolumn{4}{c}{\textbf{Content Filling}} \\
\cmidrule(lr){2-5} \cmidrule(lr){6-9} \cmidrule(lr){10-13}
 & Clean & Clean@32 & Union & Union@32 
 & Clean & Clean@32 & Union & Union@32 
 & Clean & Clean@32 & Union & Union@32 \\
\midrule
Qwen-1.7B  & 54.3 & \textbf{76.2} & 47.6 & \textit{58.1} & 37.1 & \textbf{55.2} & 27.6 & \textit{40.0} & 27.6 & \textbf{41.0} & 21.0 & \textit{30.5} \\
Qwen-8B    & 52.4 & \textbf{72.4} & 45.7 & \textit{66.7} & 38.1 & \textbf{55.2} & 30.5 & \textit{46.7} & 27.6 & \textbf{42.9} & 20.0 & \textit{31.4} \\
Qwen-32B   & 65.7 & \textbf{76.2} & 52.4 & \textit{72.4} & 38.1 & \textbf{56.2} & 30.5 & \textit{44.8} & 25.7 & \textbf{42.9} & 21.0 & \textit{31.4} \\
\bottomrule
\end{tabular}
}
\caption{Performance of Qwen models on RoTBench  with greedy decoding (Clean and Union) and {\bon} ($n=32$) with {\modelName}-14B (Clean@32 and Union@32).}
\label{tab:rotbench-results}
\end{table*}

Table~\ref{tab:rotbench-results} reports results for greedy decoding and {\bon} ($n=32$) with {\modelName}-14B. We highlight two key findings. First, {\bon} decoding yields substantial gains across all models and tasks. For instance, {\modelName} improved Qwen-8B performance on Tool Selection from 52.4 to 72.4 on the Clean split, while performance on the Union split improved from 45.7 to 66.7. Comparable gains of 15–25 points are observed for Parameter Identification and Content Filling. Second, Union@32 consistently outperforms the Clean baseline, despite Union being the more difficult split. For example, Qwen-32B achieves 72.4 on Tool Selection under Union@32 compared to 65.7 under Clean, showing that {\bon} decoding not only mitigates noise but can also exceed performance on noise-free data.

\subsection{RQ3: Reward-guided Fine-Tuning }

\subsubsection{ {\modelName} for data filtering }
\label{sec:data-filtering}

In this experiment, we assess the effectiveness of using {\modelName} as a data filter to construct a high-quality training dataset for tool-use models. We curate a training corpus comprising both single-turn and multi-turn examples drawn from APIGen-MT \citep{liu2024apigen}, SealTools \citep{wu2024sealtoolsselfinstructtoollearning}, Glaive V2\footnote{\url{https://huggingface.co/datasets/glaiveai/glaive-function-calling-v2}}, and Granite function-calling dataset \citep{abdelaziz2024granite}, yielding a total of 16K samples. We highlight that these datasets have no overlap with {\modelName} training data, thus allowing us to test the generalization capabilities of {\modelName}. We select Llama-3.1-8B-Instruct \citep{grattafiori2024llama} as the base model and performed  LoRA-based fine-tuning \citep{hu2022lora} to train each variant for 1 epoch with a learning rate of 2e-4, a LoRA rank of 16, alpha of 32, a cosine scheduler, and a warmup ratio of 10\%.

\begin{table*}[h]
\centering
\resizebox{0.85\textwidth}{!}{%
\begin{tabular}{cccccccc}
\toprule
Llama-3.1-8B-Instruct & BFCL V3 & ToolAlpaca & Nexus & API-Bank-1 & API-Bank-2 & Sealtools & Average \\
\midrule
 Base     & 49.6 & 38.0 & 64.8 & \textbf{67.9} & \textbf{66.2} & 37.6 & 54.0 \\                                                      
FT-16K   & 54.1 & 43.0 & \textbf{75.5} & 57.4 & 63.5 & 72.7 & 61.0 \\
FT-Random-8K & 55.2 & \textbf{44.0} & 74.2 & 49.9 & 54.1 & 73.2 & 58.4 \\
FT-Best-8K   & \textbf{55.4} & \textbf{44.0} & 72.0 & 63.7 & \textbf{66.2} & \textbf{73.7} & \textbf{62.5} \\
\bottomrule
\end{tabular}
}
\caption{Finetuning results of Llama-3.1-8B-Instruct on three training subsets: full 16K dataset (FT-16K), 8K randomly sampled (FT-Random-8K), and top 8K selected by {\modelName}-14B (FT-Best-8K).}
\label{data_filtering_tab}
\vspace*{-5pt}
\end{table*}

Table \ref{data_filtering_tab} compares the performance of the base model with three fine-tuned variants: (1) trained on the full 16K dataset (FT-16K), (2) trained on a random 8K subset (FT-Random-8K), and (3) trained on the top 8K samples as ranked by {\modelName}-14B (FT-Best-8K). We highlight the following key insights from the results.

\begin{itemize}[topsep=0pt, parsep=0pt, leftmargin=*]
    \item \textbf{Fine-tuning improves performance, but naive subsampling degrades it.} 
    All fine-tuned models outperform the base model, increasing accuracy from 54.0\% to 61.0\% when trained on the full dataset. 
    However, training on a random 8K subset drops accuracy to 58.4\%, showing that naive subsampling includes low-quality samples and discards high-quality ones.

    \item \textbf{{\modelName}-based data filtering achieves the best results.}
    Selecting the top 50\% of samples using {\modelName}-14B yields 62.5\% accuracy -- surpassing the full-data model while using only half the corpus -- demonstrating {\modelName}'s ability to identify high-quality data, enabling superior performance under tighter training budgets.
\end{itemize}

These results highlight the importance of data quality in fine-tuning tool-use models and show that reward-guided filtering of low-quality data can yield superior performance with less training.

\subsubsection{Policy Optimization Using {\modelName}}
\label{sec:policy-optimization}

To assess the utility of {\modelName} for policy optimization, we follow ToolRL \citep{qian2025toolrlrewardtoollearning} and train models using Group Relative Policy Optimization (GRPO) \citep{shao2024deepseekmath}. ToolRL defines the reward as $R = R_\text{format} + R_\text{correctness}$, where $R_{\text{format}} \in \{0, 1\}$ evaluates whether the output adheres to the format specified in the prompt, and $R_{\text{correctness}}$ measures the correctness of the tool-call output. We consider three variants of $R_{\text{correctness}}$:  
(1) $R_{\text{schema}} \in \{-1, 1\}$, validates the predicted tool calls against the provided tool specifications, assigning $-1$ if there are schema violations and $+1$ otherwise;  
(2) $R_{\text{ToolRL}} \in [-3, 3]$, follows \cite{qian2025toolrlrewardtoollearning} and computes rewards by comparing predicted and ground-truth tool calls; and
(3) $R_{\text{ToolRM}} \in [-3, 3]$, scores the tool calls using {\modelName}-14B. Notably, $R_\text{schema}$ and $R_\text{ToolRM}$ do not require access to ground-truth tool calls, making them more appropriate for RL settings, whereas $R_\text{ToolRL}$ requires ground truth, limiting its applicability.

\begin{table}[]
    \centering
    \resizebox{0.85\linewidth}{!}{%
    \begin{tabular}{cccc}
        \toprule
        \textbf{Model}                                  &  \textbf{\makecell{Reward\\Variant}}    & \textbf{\makecell{Non-Live \\ AST Acc}} & \textbf{\makecell{Live AST \\ Acc}} \\
        \midrule
        \multirow{4}{*}{\makecell{Llama-3.2-\\3B-Instruct}} & Base & 15.35\%           & 43.82\%      \\
                                       & $R_\text{Schema}$   & 51.71\%          & 62.25\%      \\
                                       & $R_\text{ToolRL}$ & \textit{75.27}\%          & \textit{64.25}\%     \\
                                       & $R_\text{ToolRM}$   & \textbf{78.40}\%          & \textbf{64.32}\%      \\
        \midrule                                       
        \multirow{4}{*}{\makecell{Qwen2.5-\\3B-Instruct}}   & Base & 43.06\%          & 55.66\%      \\
                                               & $R_\text{Schema}$   & 63.17\%          & 66.54\%      \\
                                               & $R_\text{ToolRL}$ & \textbf{80.42}\%          & \textit{67.21}\%      \\
                                               & $R_\text{ToolRM}$   & \textit{79.58}\%          & \textbf{67.51}\%      \\
        \bottomrule
                                               
        \end{tabular}
        }
    \caption{BFCL-v3 performance of Llama-3.2-3B-Instruct and Qwen2.5-3B-Instruct trained with GRPO under three reward designs.}
    \label{tab:grpo-results}
    \vspace{-10pt}
\end{table}

We train Llama-3.2-3B-Instruct and Qwen2.5-3B-Instruct models using the three reward variants and evaluate them on the BFCL-v3 dataset, with results shown in Table \ref{tab:grpo-results}. Across both models, all three reward variants substantially improve performance over the base models. Simple schema-based rewards provide strong gains without requiring ground-truth supervision, yielding average improvements of about 20 points. 
{\modelName}-based rewards achieve the best overall results: for Llama-3.2-3B-Instruct, $R_{\text{ToolRM}}$ attains the highest accuracy on both evaluation metrics, and for Qwen2.5-3B-Instruct, it provides the best Live AST accuracy, surpassing the gold-dependent $R_{\text{ToolRL}}$. Overall, $R_{\text{ToolRM}}$ consistently matches or exceeds $R_{\text{ToolRL}}$ despite not requiring access to ground truth, highlighting its practicality and effectiveness as a scalable reward signal for reinforcement learning in the tool-calling setting. Additional experimental details are provided in Appendix \ref{apdx:grpo-details}.

%% file: sections_acl/conclusion.tex
In this paper, we presented a comprehensive framework for reward modeling in tool-calling scenarios. Our benchmark, FC-RewardBench, enables systematic assessment of reward models on tool-calling tasks. We also presented a framework for training outcome RMs that outperform existing significantly larger RMs in the tool calling setting. 
Overall, {\modelName} enables effective inference-time scaling, data-efficient fine-tuning, and ground-truth–free policy optimization.
Looking ahead, we see several promising directions for advancing reward modeling in this domain. First, moving beyond classification-based RMs to generative verifiers with chain-of-thought reasoning could improve robustness and interpretability. Second, incorporating the tool and environment state into training could help models safely recover from execution failures. Finally, bridging outcome and process reward modeling may offer a unified framework that balances scalability with fine-grained control over reasoning quality.

%% file: sections_acl/appendix.tex
\lstset{
  basicstyle=\normalsize,
  columns=fullflexible,
  frame=single,
  breaklines=true,
  numbers=left,
  postbreak=\mbox{\textcolor{red}{$\hookrightarrow$}\space},
}

\begin{table*}[t]
\centering
\resizebox{0.85\textwidth}{!}{%
\begin{tabular}{cccccccc}
\toprule
Dataset & \makecell{\# Examples} & \makecell{\# Tools \\ (avg./query)} & \makecell{\# MT \\ queries} & \makecell{Avg. MT \\ turns} & \makecell{Nested \\ calls} & \makecell{Avg. output \\ tool calls} & \makecell{Data \\ source} \\
\midrule
BFCL-v3      & 4,441 & 2,631 (3.3) & 800 & 4.2 & \checkmark & 2.4 & Real \\
API-Bank     &   473 &    64 (3.4) & 397 & 3.4 & \text{\sffamily x} & 1.0 & Real \\
ToolAlpaca   &   100 &    64 (5.6) &   0 & –   & \text{\sffamily x} & 1.5 & Synthetic \\
NexusRaven   &   318 &    65 (7.4) &   0 & –   & \text{\sffamily x} & 1.0 & Synthetic \\
SealTools    &   627 & 3,036 (9.9) &   0 & –   & \checkmark & 2.9 & Synthetic \\
\bottomrule
\end{tabular}%
}
\caption{Statistics of the evaluation benchmarks. ``MT'' denotes multi-turn queries.}
\label{tab:eval-benchmarks}
\end{table*}

\subsection{{\fcrewardbench} benchmark details}
\label{apdx:fcrb-benchmark-details}

The list of models included in {\fcrewardbench}, along with the number of incorrect tool call output samples per model is provided in Table \ref{tab:fcrb-model-breakdown}.

\begin{table}[h]
\centering
\begin{tabular}{lc}
\toprule
Model Name                              & Count \\
\midrule
Qwen/Qwen2.5-0.5B-Instruct              & 450   \\
Qwen/Qwen2.5-0.5B-Instruct-FC           & 237   \\
ibm-granite/granite-20b-functioncalling & 112   \\
Qwen/Qwen2.5-1.5B-Instruct              & 102   \\
BitAgent/BitAgent-8B                    & 74    \\
DeepSeek-R1                             & 64    \\
openbmb/MiniCPM3-4B-FC                  & 59    \\
NovaSky-AI/Sky-T1-32B-Preview           & 54    \\
Qwen/Qwen2.5-1.5B-Instruct-FC           & 52    \\
speakleash/Bielik-11B-v2.3-Instruct     & 41    \\
Qwen/Qwen2.5-14B-Instruct-FC            & 38    \\
openbmb/MiniCPM3-4B                     & 38    \\
Qwen/Qwen2.5-14B-Instruct               & 28    \\
Qwen/Qwen2.5-7B-Instruct                & 23    \\
ZJared/Haha-7B                          & 22    \\
meetkai/functionary-small-v3.1-FC       & 21    \\
watt-ai/watt-tool-70B                   & 21    \\
Qwen/Qwen2.5-7B-Instruct-FC             & 18    \\
Qwen/Qwen2.5-32B-Instruct-FC            & 15    \\
Qwen/Qwen2.5-32B-Instruct               & 13    \\
meetkai/functionary-medium-v3.1-FC      & 11    \\
Team-ACE/ToolACE-2-8B                   & 6     \\
Qwen/QwQ-32B-Preview                    & 1    \\
\bottomrule
\end{tabular}
\caption{Breakdown of errors by models in {\fcrewardbench}}
\label{tab:fcrb-model-breakdown}
\end{table}

\begin{figure}[h]
    \centering
    \includegraphics[width=\linewidth]{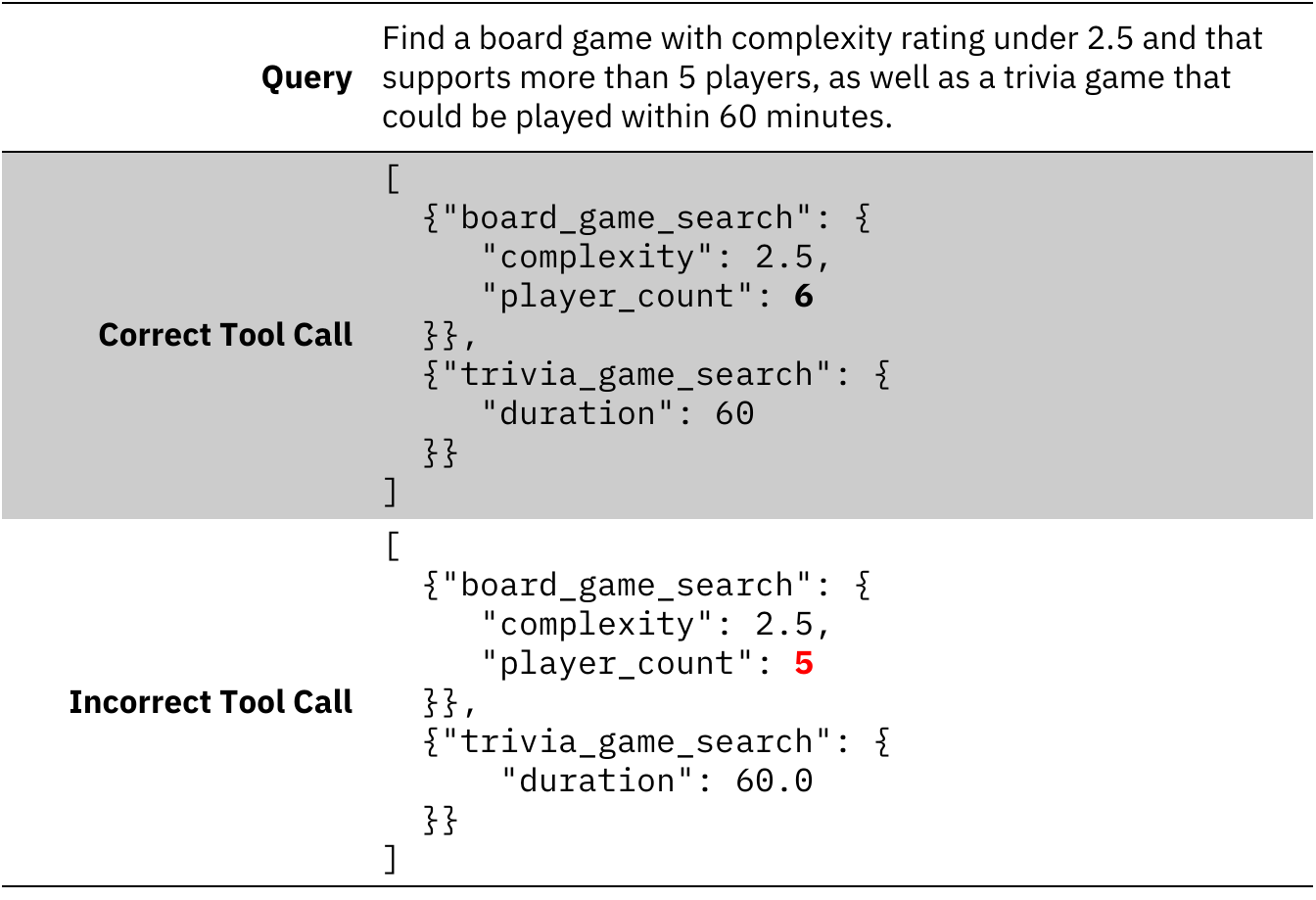}
    \caption{Representative example from \fcrewardbench\; the parameter \texttt{player\_count} is set to an incorrect value. The tool catalog is hidden for brevity.}
    \label{fig:fc-reward-bench-example}
\end{figure}

A representative example from the {\fcrewardbench} dataset is shown in Figure \ref{fig:fc-reward-bench-example}.

\subsection{{\modelName} Training Data Details}
\label{apdx:training-datagen-details}

We use the following models to generate the training data for {\modelName}:

{\scriptsize
\begin{itemize}
    \item \texttt{\href{https://huggingface.co/ibm-granite/granite-3.3-2b-instruct}{ibm-granite/granite-3.3-2b-instruct}}
    \item \texttt{\href{https://huggingface.co/ibm-granite/granite-3.3-8b-instruct}{ibm-granite/granite-3.3-8b-instruct}}
    \item \texttt{\href{https://huggingface.co/ibm-granite/granite-20b-functioncalling}{ibm-granite/granite-20b-functioncalling}}
    \item \texttt{\href{https://huggingface.co/HuggingFaceTB/SmolLM2-1.7B-Instruct}{HuggingFaceTB/SmolLM2-1.7B-Instruct}}
    \item \texttt{\href{https://huggingface.co/Qwen/Qwen2.5-0.5B-Instruct}{Qwen/Qwen2.5-0.5B-Instruct}}
    \item \texttt{\href{https://huggingface.co/Qwen/Qwen2.5-1.5B-Instruct}{Qwen/Qwen2.5-1.5B-Instruct}}
    \item \texttt{\href{https://huggingface.co/Qwen/Qwen2.5-7B-Instruct}{Qwen/Qwen2.5-7B-Instruct}}
    \item \texttt{\href{https://huggingface.co/Qwen/Qwen2.5-14B-Instruct}{Qwen/Qwen2.5-14B-Instruct}}
    \item \texttt{\href{https://huggingface.co/Qwen/Qwen2.5-32B-Instruct}{Qwen/Qwen2.5-32B-Instruct}}
    \item \texttt{\href{https://huggingface.co/mistralai/Mistral-7B-Instruct-v0.3}{mistralai/Mistral-7B-Instruct-v0.3}}
    \item \texttt{\href{https://huggingface.co/mistralai/Mistral-Nemo-Instruct-2407}{mistralai/Mistral-Nemo-Instruct-2407}}
\end{itemize}
}

\begin{figure*}
    \centering
    \includegraphics[width=0.9\textwidth]{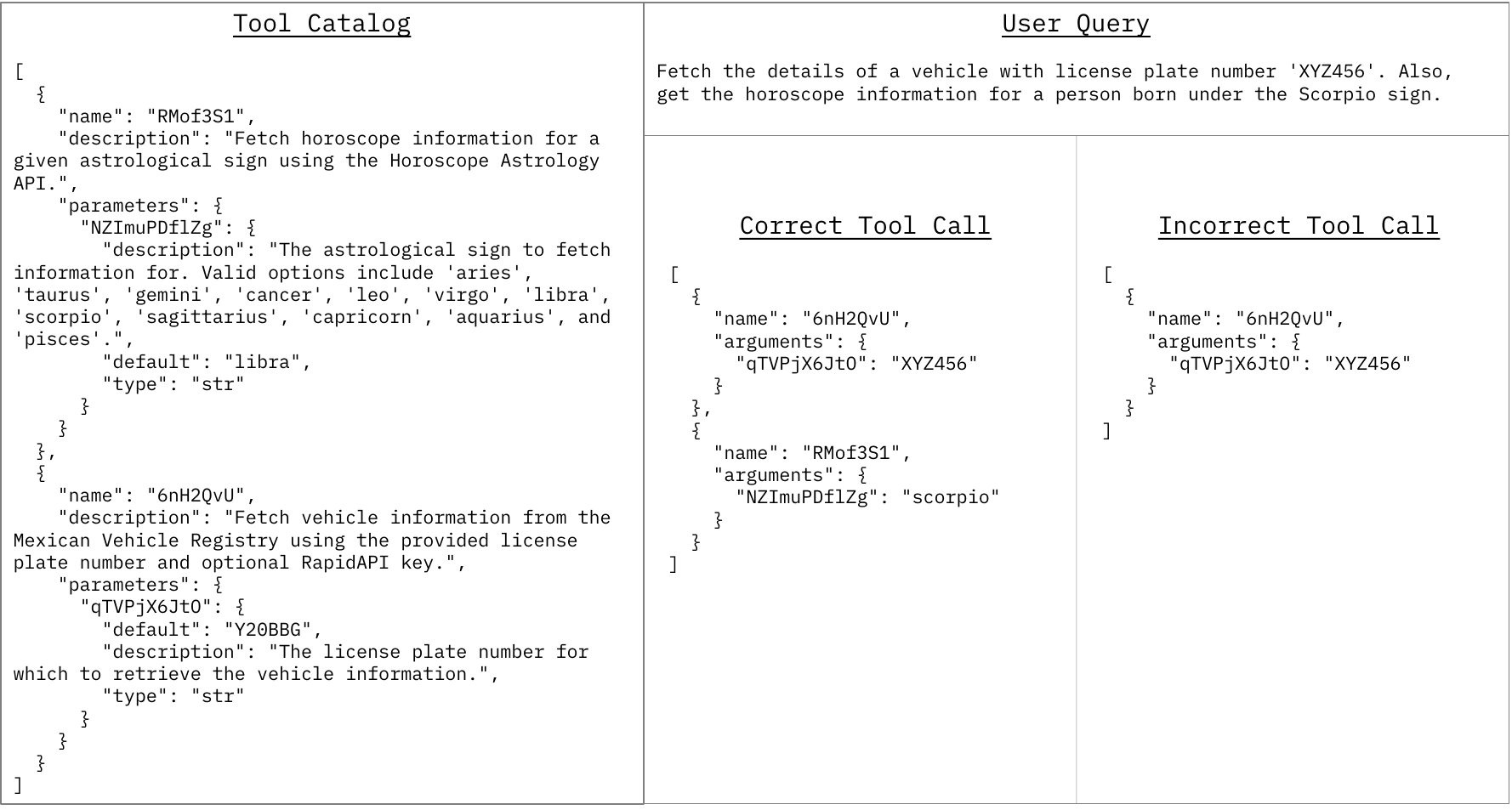}
    \\
    \vspace{10pt}
    \includegraphics[width=0.9\textwidth]{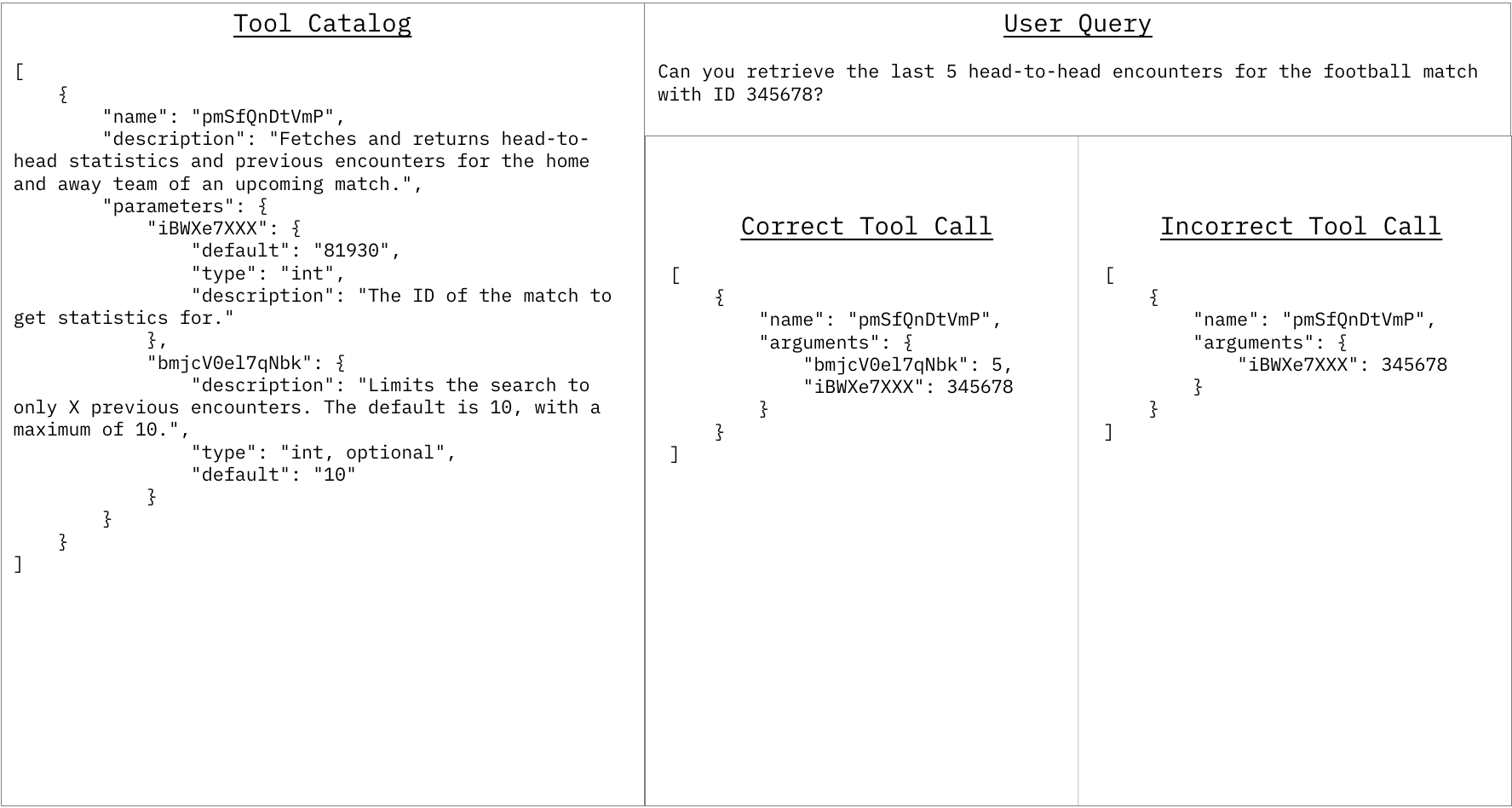}
    \caption{Data samples from {\modelName} training data. Each sample has a tool catalog, a conversation between the user and assistant, along with the corresponding correct and incorrect tool calls. The top sample is missing one tool call from the Incorrect version, while the Bottom sample is missing a parameter from the tool call.}
    \label{fig:training-data-samples}
\end{figure*}

A few samples from the training data are shown in Figure \ref{fig:training-data-samples}.

\subsection{{\modelName} prompt}
\label{apdx:toolrm-prompt}
The prompt used to train the {\modelName} is shown in Listing \ref{rmprompt}.

{
\scriptsize
\begin{lstlisting}[float=*, caption={\modelName} prompt, captionpos=b, label=rmprompt]
<|im_start|>system
You are provided with a user query, a catalog of tools available to fulfill that user query, and a list of tool calls that use tools available in the catalog to fulfill user request.
Your job is to assess whether the tool calls adequately fulfill the user request or not.

You have the following tools available:

```json
[
        {"name": "diabetes_prediction", "description": "Predict the likelihood of diabetes type 2 based on a person's weight and height.", "parameters": {"type": "dict", "properties": {"weight": {"type": "integer", "description": "Weight of the person in lbs."}, "height": {"type": "integer", "description": "Height of the person in inches."}, "activity_level": {"type": "string", "enum": ["sedentary", "lightly active", "moderately active", "very active", "extra active"], "description": "Physical activity level of the person."}}, "required": ["weight", "height", "activity_level"]}}
]
```

<|im_end|>
<|im_start|>user
Predict whether a person with weight 150lbs and height 5ft 10in who is lightly active will get type 2 diabetes.<|im_end|>
<|im_start|>assistant
```json
[
        {"diabetes_prediction": {"weight": 150, "height": 68, "activity_level": "lightly active"}}
]
```<|im_end|>
\end{lstlisting}
}

\subsection{{\fcrewardbench} Experiment Details}
\label{apdx:fcrb-experiment-details}

\subsubsection{Model Details}

We include the following models from RewardBench:

{\scriptsize
\begin{itemize}
    \item \texttt{\href{https://huggingface.co/Ray2333/GRM-Llama3.2-3B-rewardmodel-ft}{Ray2333/GRM-Llama3.2-3B-rewardmodel-ft}}
    \item \texttt{\href{https://huggingface.co/Skywork/Skywork-Reward-V2-Qwen3-4B}{Skywork/Skywork-Reward-V2-Qwen3-4B}}
    \item \texttt{\href{https://huggingface.co/Skywork/Skywork-Reward-V2-Qwen3-8B}{Skywork/Skywork-Reward-V2-Qwen3-8B}}
    \item \texttt{\href{https://huggingface.co/LxzGordon/URM-LLaMa-3.1-8B}{LxzGordon/URM-LLaMa-3.1-8B}}
    \item \texttt{\href{https://huggingface.co/Skywork/Skywork-Reward-Llama-3.1-8B-v0.2}{Skywork/Skywork-Reward-Llama-3.1-8B-v0.2}}
    \item \texttt{\href{https://huggingface.co/nicolinho/QRM-Llama3.1-8B-v2}{nicolinho/QRM-Llama3.1-8B-v2}}
    \item \texttt{\href{https://huggingface.co/infly/INF-ORM-Llama3.1-70B}{infly/INF-ORM-Llama3.1-70B}}
    \item \texttt{\href{https://huggingface.co/Skywork/Skywork-Critic-Llama-3.1-70B}{Skywork/Skywork-Critic-Llama-3.1-70B}}
\end{itemize}
}

We include the following LLMs as Judges:
{\scriptsize
\begin{itemize}
    \item \texttt{\href{https://huggingface.co/deepseek-ai/DeepSeek-V3}{deepseek-ai/DeepSeek-V3}}
    \item \texttt{\href{https://huggingface.co/meta-llama/Llama-3.1-405B-Instruct-FP8}{meta-llama/Llama-3.1-405B-Instruct-FP8}}
    \item \texttt{\href{https://huggingface.co/meta-llama/Llama-3.3-70B-Instruct}{meta-llama/Llama-3.3-70B-Instruct}}
    \item \texttt{\href{https://huggingface.co/meta-llama/Llama-4-Scout-17B-16E}{meta-llama/Llama-4-Scout-17B-16E}}
    \item \texttt{\href{https://huggingface.co/meta-llama/Llama-4-Maverick-17B-128E-Instruct}{meta-llama/Llama-4-Maverick-17B-128E-Instruct}}
    \item \texttt{\href{https://huggingface.co/openai/gpt-oss-120b}{openai/gpt-oss-120b}}
\end{itemize}
}

Additionally, we include the Tool-Augmented Reward Model (\texttt{\href{https://huggingface.co/ernie-research/Themis-7b}{ernie-research/Themis-7b}}).

\subsubsection{LLM-as-Judge Prompt}

The prompt used to evaluate LLMs-as-Judges on {\fcrewardbench} is shown in Listing \ref{llmasjudgeprompt}. The placeholders \texttt{\{tool-library\}}, \texttt{\{query\}}, \texttt{\{response-A\}}, and \texttt{\{response-B\}} are replaced with appropriate values.

{
\scriptsize
\begin{lstlisting}[float=*, caption=LLM-as-Judge Prompt used for {\fcrewardbench} evaluation, captionpos=b, label=llmasjudgeprompt]
Please act as an impartial judge and evaluate the quality of the responses provided by two AI assistants to the user question displayed below. You should choose the assistant that follows the user's instructions and answers the user's question best. 

You will be given:
- TOOL SPECIFICATIONS: All the tool specifications including parameters and their descriptions available to the assistant to answer the query.
- CONVERSATION: Conversation between the user and the assistant
- ASSISTANT RESPONSES: List of responses from two assistants - [[A]] and [[B]]. Each response is a sequence of tool calls needed to answer the question.

When comparing two tool call sequences for the same user query, carefully evaluate both sequences and determine which one better follows the tool specifications and the question requirements.

Consider the following instructions to compare the assistant responses:
- Check whether all the tools used are relevant and actually exist.
- Verify that the tools are called in a correct and logical order.
- Ensure that the correct parameters are used for each tool and that no nonexistent parameters are included.
- Confirm that parameter values and formats are appropriate based on the question and tool specifications.
- Make sure all required parameters and data mentioned in the question are included.
- Look for any extra tools or parameters that are not needed or mentioned in the question.

Also, follow these general instructions:
- Begin your evaluation by comparing the two responses (i.e., [[A]] and [[B]]) and provide a short explanation.
- Avoid any position biases and ensure that the order in which the responses were presented does not influence your decision.
- Do not allow the length of the responses to influence your evaluation.
- Do not favor certain names of the assistants.
- Be as objective as possible.
- After providing your explanation, output your final verdict by strictly following this format: "[[A]]" if assistant A's response is better and "[[B]]" if assistant B's response is better.
- STRICTLY follow this output format: [EXPLANATION]\n{Short comparison highlighting differences and reasoning.}\n\n[VERDICT]\n{[[A]] or [[B]]}. Place your reasoning after the [EXPLANATION] tag and your final choice after the [VERDICT] tag.

# TOOL SPECIFICATIONS 
{tool-library}

# CONVERSATION
{query}

# ASSISTANT RESPONSES: 
[[A]] {response-A}
[[B]] {response-B}

# ANSWER:
\end{lstlisting}
}

\subsection{Correlation with performance on downstream tasks:} 
\label{apdx:fcrb-correlation}

\begin{figure}
    \begin{center}
        \includegraphics[width=0.95\linewidth]{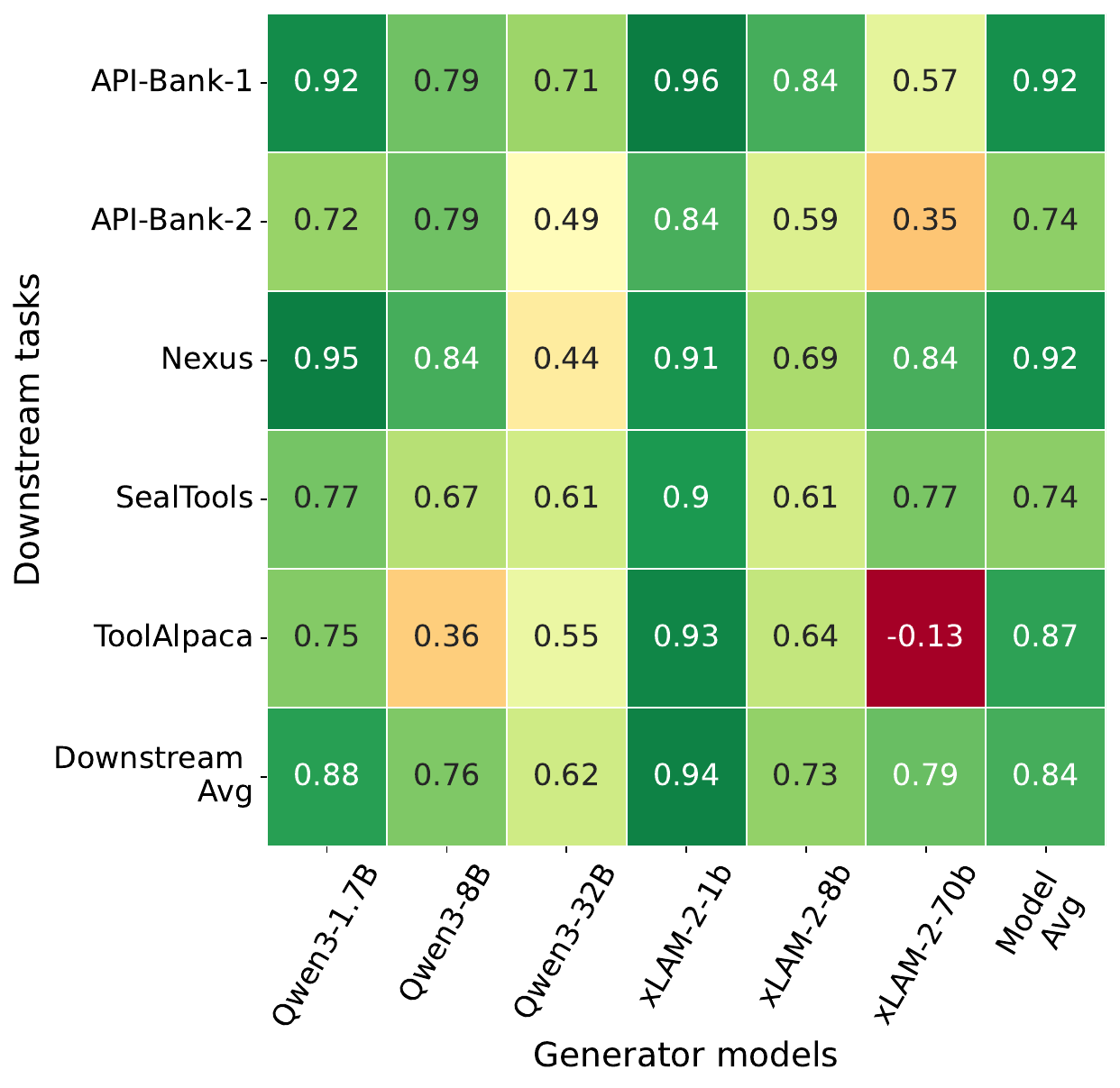}
    \end{center}
    \caption{Correlation heatmap between {\fcrewardbench} performance and downstream accuracy across generator models and benchmarks, showing consistently strong alignment (avg.\ correlation = 0.84).}
    \label{fig:fcrb-correlation}
\end{figure}

The primary purpose of \fcrewardbench\ is to enable quick evaluation of RMs without having to do computationally expensive downstream evaluation. It is thus imperative that performance on {\fcrewardbench} reflects downstream task performance. 
To assess this, we select six generator models (Qwen3-1.7B, 8B, 32B, and xLAM-1B, 8b, 70B), 11 RMs (eight RMs from RewardBench and three {\modelName} variants), and five benchmarks. For each generator model, RM, and dataset combination, we compute the performance in a {\bon} $(n=32)$ setting and compute the Pearson correlation coefficient between the {\bon} performance and RM performance on {\fcrewardbench}. Results are shown in Figure \ref{fig:fcrb-correlation}.

Overall, we find that {\fcrewardbench} scores are strongly correlated with downstream task accuracy, with an average correlation of 0.84 across benchmarks and generator models. Across generator models, the average correlation ranges from $0.62$ to $0.94$, indicating that the alignment between {\fcrewardbench} and downstream performance is robust across model families. Importantly, this correlation remains stable even at scale: larger models such as Qwen3-32B and xLAM-2-70B continue to exhibit strong agreement between {\fcrewardbench} accuracy and downstream results. Taken together, these findings confirm that {\fcrewardbench} provides a reliable and computationally efficient proxy for expensive downstream evaluations.

\subsection{Error Analysis}
\label{apdx:bon-analysis}

In this experiment, we evaluate the impact of {\modelName}-14B on error reduction in the {\bon} $(n=32)$ sampling setting, using the Qwen3-1.7B generator on the single-turn splits of BFCL-v3. As reported in Table~\ref{tab:error_breakdown}, {\modelName}-14B decreases the total error count from 742 to 573, corresponding to a 22.7\% relative reduction. The predominant error type, Incorrect Parameter Value, responsible for nearly 42\% of greedy decoding errors, is reduced by 28\%, indicating that {\modelName} is particularly effective at mitigating semantic mis-specification of parameter values. Moreover, errors such as Incorrect Function Name and Wrong Number of Functions are reduced by 37\% and 57\%, respectively. In contrast, Irrelevance Errors increase from 185 to 210, suggesting that {\modelName} tends to favor producing function call outputs even in cases where no valid call can appropriately satisfy the user query.

\begin{table}[t!]
\centering\
\resizebox{0.95\linewidth}{!}{%
\begin{tabular}{lcc}
\toprule
Error type                 & Greedy & {\modelName}-14B \\
\midrule
Incorrect Parameter Value  & 321    & 231    \\
Irrelevance error          & 185    & 210    \\
Malformed output syntax    & 86     & 34     \\
Incorrect function name    & 62     & 39     \\
Missing optional parameter & 36     & 26     \\
Incorrect parameter type   & 31     & 24     \\
Wrong number of functions  & 21     & 9      \\
\midrule
Total                      & 742    & 573   \\
\bottomrule
\end{tabular}
}
\caption{Breakdown of errors with the Qwen3-1.7B model as generator with Greedy decoding and Best-of-32 sampling with {\modelName}-14B}
\label{tab:error_breakdown}
\end{table}

\subsection{{\modelName} generalization}

\begin{figure}
    \centering
    \includegraphics[width=\linewidth]{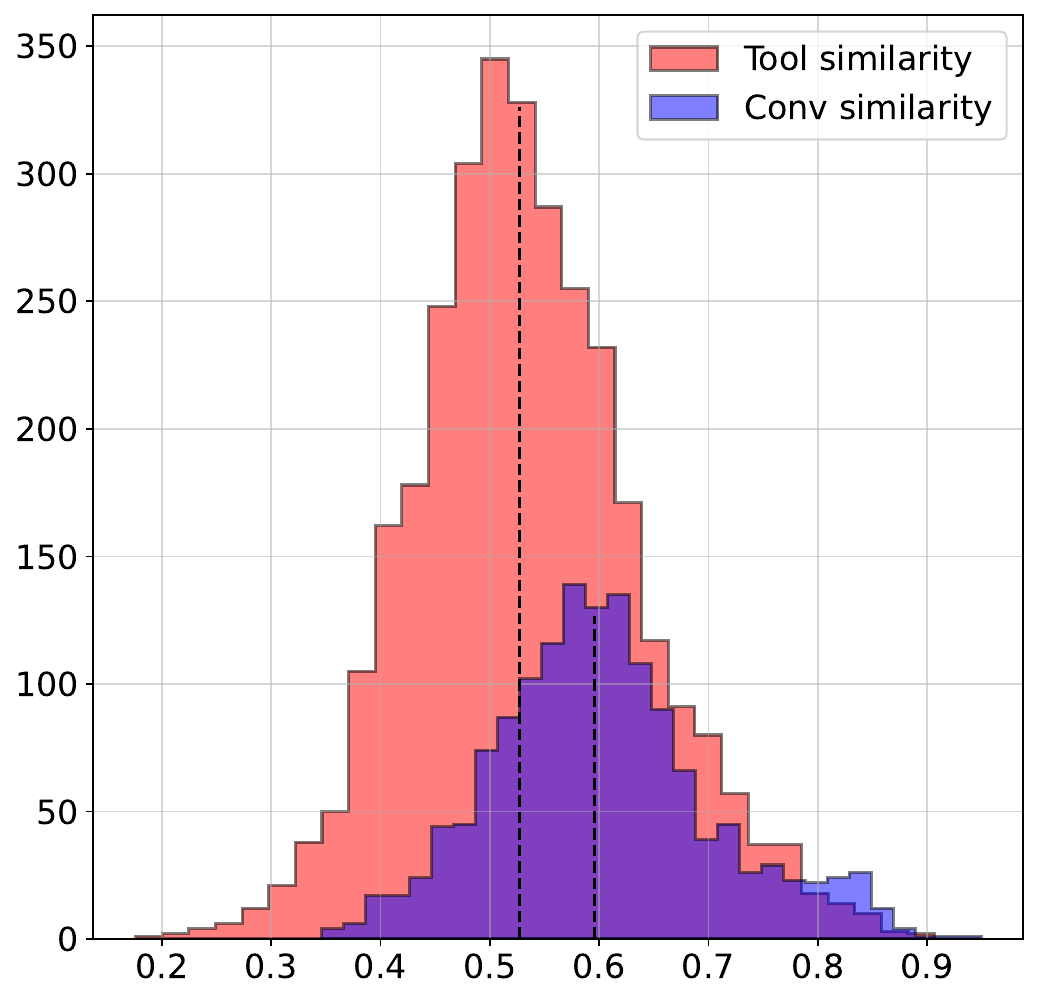}
    \caption{Histogram of maximum cosine similarity between train and test tools and conversations.}
    \label{fig:toolrm-generalizability}
\end{figure}

To assess how well {\modelName} generalizes to inputs unseen during training, we embed each tool and conversation from both the training set and the five non-BFCL test sets using the \texttt{all-mpnet-base-v2}\footnote{https://huggingface.co/sentence-transformers/all-mpnet-base-v2} embedding model. We then compute, for each test example, the maximum cosine similarity with the training set and plot the resulting distributions in Figure~\ref{fig:toolrm-generalizability}. We observe low similarity between training and test examples for both tools and conversations (median $<$ 0.6 in both cases), indicating that {\modelName} generalizes effectively to novel inputs at test time.

\subsection{GRPO Experiment Details}
\label{apdx:grpo-details}

We utilize the open-source library \texttt{verl}\footnote{https://github.com/volcengine/verl} and the code base and dataset provided by ToolRL~\footnote{https://github.com/qiancheng0/ToolRL} to train models for GRPO experiments. The hyperparameters used to train the model are listed in Table \ref{tab:grpo-hyperparams}, and the prompt used to train the models is listed in Listing \ref{toolrl_prompt}.

\begin{table}[h]
\centering
\resizebox{0.6\linewidth}{!}{%
\begin{tabular}{cc}
\toprule
\textbf{Hyperparameter}      & \textbf{Value} \\
\midrule
Max prompt length   & 2048  \\
Max response length & 1024  \\ 
Learning rate       & 1e-6  \\
PPO mini batch size & 128   \\
Number of rollouts    & 4 \\
Total epochs          & 15    \\
Train batch size      & 512   \\
Validation batch size & 128  \\
\bottomrule
\end{tabular}%
}
\caption{GRPO Experiment Hyperparameters}
\label{tab:grpo-hyperparams}
\end{table}

{
\scriptsize
\begin{lstlisting}[float=*, caption=Prompt used for GRPO training of models, captionpos=b, label=toolrl_prompt]
You are a helpful multi-turn dialogue assistant capable of leveraging tool calls to solve user tasks and provide structured chat responses.

**Available Tools**
```json
{{TOOLS}}
```

**Steps for Each Turn**
1. **Think:** Recall relevant context and analyze the current user goal.
2. **Decide on Tool Usage:** If a tool is needed, specify the tool and its parameters.
3. **Respond Appropriately:** If a response is needed, generate one while maintaining consistency across user queries.

**Output Format**
```plaintext
<think> Your thoughts and reasoning </think>
<tool_call>
{"name": "Tool name", "parameters": {"Parameter name": "Parameter content", "... ...": "... ..."}}
{"name": "... ...", "parameters": {"... ...": "... ...", "... ...": "... ..."}}
...
</tool_call>
<response> AI's final response </response>
```

**Important Notes**
1. You must always include the `<think>` field to outline your reasoning. Provide at least one of `<tool_call>` or `<response>`. Decide whether to use `<tool_call>` (possibly multiple times), `<response>`, or both.
2. You can invoke multiple tool calls simultaneously in the `<tool_call>` fields. Each tool call should be a JSON object with a "name" field and an "parameters" field containing a dictionary of parameters. If no parameters are needed, leave the "parameters" field an empty dictionary.
3. Refer to the previous dialogue records in the history, including the user's queries, previous `<tool_call>`, `<response>`, and any tool feedback noted as `<obs>` (if exists).
\end{lstlisting}
}

\subsection{Ablation analysis}
\label{sec:ablation}

We conduct an ablation study to assess the contribution of individual components to the overall performance of {\modelName}. Specifically, we train {\modelName}-1.5B, ablating different hyperparameters, training datasets, and generator models, and evaluating each variant on the {\fcrewardbench} dataset. Table \ref{tab:ablation-exp} summarizes the results. 
First, we observe that the full {\modelName}-1.5B model, incorporating all components, achieves the highest performance. Second, removing the API-Gen dataset leads to a 16.4-point drop in performance, underscoring its significance in training. Finally, obfuscating tool and parameter names results in a 13.63-point reduction in performance. This suggests that obfuscation prevents the model from overfitting to specific tool or parameter names and encourages it to attend to other parts of the tool specifications, thereby improving robustness and generalization.

\begin{table}[h!]
    \centering
    \resizebox{0.85\linewidth}{!}{%
    \begin{tabular}{c c}
          \toprule
       \textbf{Model}  & \textbf{Accuracy} \\
       \midrule
       
       {\modelName}-1.5B  & 81.88\% \\
       \midrule
       \multicolumn{2}{c}{\textbf{Hyperparameter Ablation}}    \\
       \midrule
       Without obfuscation      &    68.25\%    \\
       High reward centering ($\eta=0.1$)    &   80.02\% \\
       No reward centering ($\eta=0$)      &   80.32\%    \\
       \midrule
       \multicolumn{2}{c}{\textbf{Data Ablation}}    \\
       \midrule
       Without API-Gen             &    65.48\%    \\
       Without SGD              &    81.03\%    \\
       \midrule
       \multicolumn{2}{c}{\textbf{Generator model ablation}}    \\
       \midrule
       Only large ($>=$12B) models    &    78.70\%    \\
       Only small ($<=$2B) models    &    81.01\%    \\

       \bottomrule
    \end{tabular}
    }
    \caption{Ablation results for {\modelName}-1.5B model on {\fcrewardbench} dataset.}
    \label{tab:ablation-exp}
\end{table}